# Railway Track Specific Traffic Signal Selection Using Deep Learning

S. Ritika, Shruti Mittal, Dattaraj Rao

General Electric

## ABSTRACT

*With the railway transportation Industry moving actively towards automation, accurate location and inventory of wayside track assets like traffic signals, crossings, switches, mileposts, etc. is of extreme importance. With the new Positive Train Control (PTC) regulation coming into effect, many railway safety rules will be tied directly to location of assets like mileposts and signals. Newer speed regulations will be enforced based on location of the Train with respect to a wayside asset. Hence it is essential for the railroads to have an accurate database of the types and locations of these assets.*

*This paper talks about a real-world use-case of detecting railway signals from a camera mounted on a moving locomotive and tracking their locations. The camera is engineered to withstand the environment factors on a moving train and provide a consistent steady image at around 30 frames per second. Using advanced image analysis and deep learning techniques, signals are detected in these camera images and a database of their locations is created.*

*Railway signals differ a lot from road signals in terms of shapes and rules for placement with respect to track. Due to space constraint and traffic densities in urban areas signals aren't placed on the same side of the track and multiple lines can run in parallel. Hence there is need to associate signal detected with the track on which the train runs. We present a method to associate the signals to the specific track they belong to using a video feed from the front facing camera mounted on the lead locomotive. A pipeline of track detection, region of interest selection, signal detection has been implemented which gives an overall accuracy of 94.7% on a route covering 150km with 247 signals.*

## 1. INTRODUCTION

Traffic signal is a very critical railway asset. Correct detection of the same is very important to envisage automated movement of locomotives as well maintaining an up to date track asset database. There have been many efforts in the past to detect the same. [1], [2] discuss conventional signal detection techniques, recently a lot of efforts have been put into application of deep learning for object detection. COCO and ILSVRC competitions have provided vast dataset which has accelerated the advancements in object detection. Models like Faster-RCNN can achieve high accuracy for a large collection of daily life objects one being traffic signal. But railway signals have some peculiarities which distinguish them from road traffic signals. There are different types of signals varying in shapes like 2 aspect, 3 aspect, shunt lights, route indicator as well as spatially like overhead signal, ground lights, on pole etc. as shown in Figure 1. Some of the shapes are very different from the ones found on road. The shunt indicator is a case very specific to railroads. Thus, the spatial and topographical variations must be learnt by the model. Additionally, for maintaining track database or performing such on-board signal detection for railway lines, there is a need to associate the signals to the specific track.

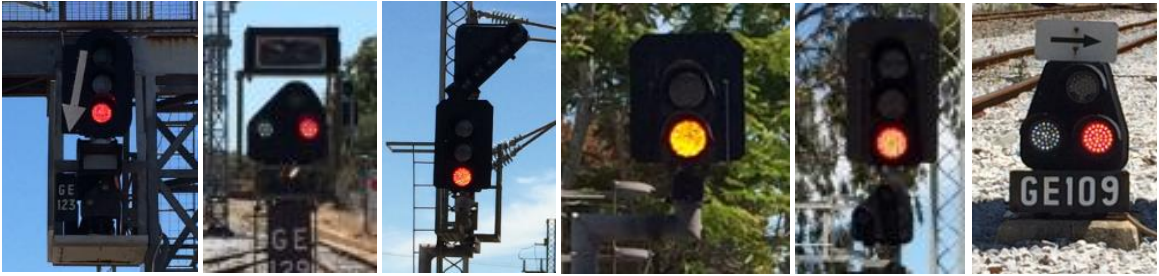

*Figure 1 Different types of rail road signals*

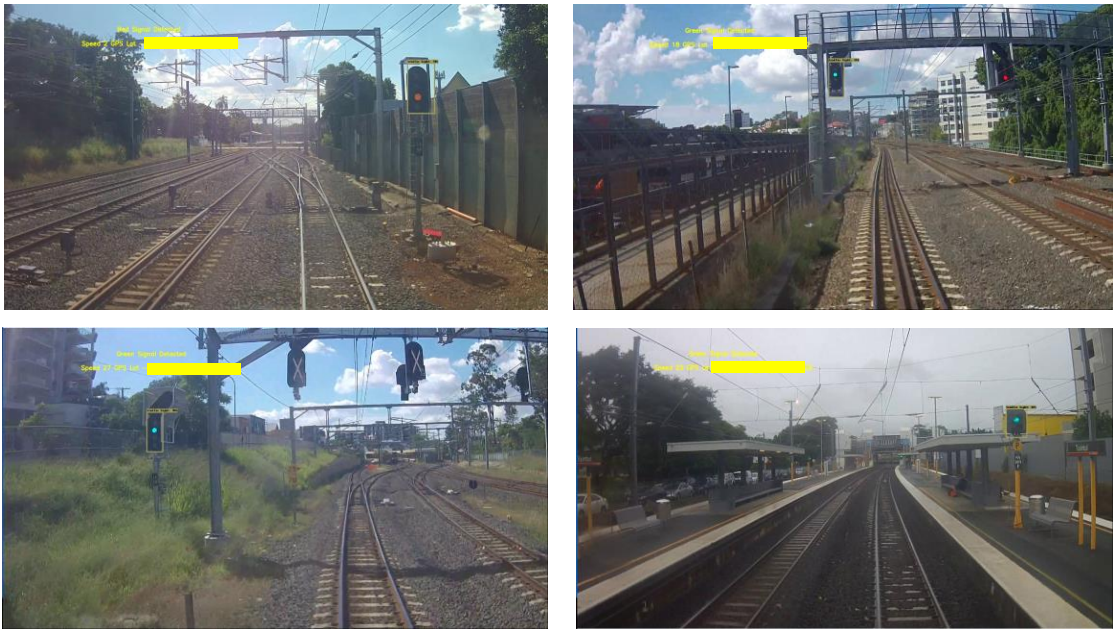

*Figure 2 Spatial and topographical variation of signal placement*

## 2. PROBLEM DESCRIPTION

The problem described in this paper is to associate the signals to the specific track they belong to. The input is video feed from the front facing camera mounted on the locomotive. Since the images obtained from the video feed lack depth information, and no detection and ranging equipment like LIDAR, ultrasonic device is available on locomotive, it is likely that the distant signals which are proximate to the current track might be associated with the same. The placement of signal with respect to track is not fixed and can vary immensely depending on the type of signal and as the locomotive moves on the track as seen in Figure 2.

Additionally, due to space constraint and rail traffic densities in urban areas, multiple lines run in parallel. Also, the signals aren't placed on the same side w.r.t. the track. It is placed mostly to the left of the track except for the rightmost track which has its signal on the right. Signals are also placed on the platform of the station rather than at the very end. The environmental conditions change the appearance drastically. These add up to the already complex task of track specific signal detection.

# 3. APPROACH

## 3.1 FIXED REGION OF INTEREST (ROI)

The simplest approach to the problem can be to select a fixed region of interest based on heuristics and run object detection algorithm on that. But it cannot take the spatial variation of signal placement into consideration as the region is fixed. If a bigger region is taken considering all possible signal placements, many signals of other track are detected, thus increasing the false positive rate.

## 3.2 FIXED ROI BASED ON RULES

Another intuitive approach is to keep checking on the left half of the image, if no signal is found check to the right. But this approach is also susceptible to failure since signals of different tracks aren't always collocated. For the specific case when the locomotive is on the rightmost track, it is bound to fail as it will keep detecting the signals of other track on the left rather than its signal placed on the right as seen in Figure 2.

## 3.3 OBJECT CLASSIFICATION BASED ON DEEP LEARNING

Another approach can be to categorize the available data and label it considering whether the signal corresponds to the current track or not. This can potentially make the algorithm learn the heuristic logic mentioned above like humans do: just by observing the video feed [3]. But the amount data required for this will be huge. Also, taking the average labelling speed of a human being to be 10 images/min it will involve a lot of labelling time. Also, a lot of advancements have occurred in the object detection domain. We might not be able to utilize these pre-trained models to the fullest and depend on the limited data we have in this approach.

## 3.4 SELECTED APPROACH

The position of signal is highly dependent on the position of the current track relative to the other tracks present. Hence, it would be wise to detect the tracks on the image and position the signals detected relative to that. Thus, the deep learning based object classification approach is divided into two steps: detecting the tracks, and relating the signals corresponding to these tracks as shown in Figure 3.

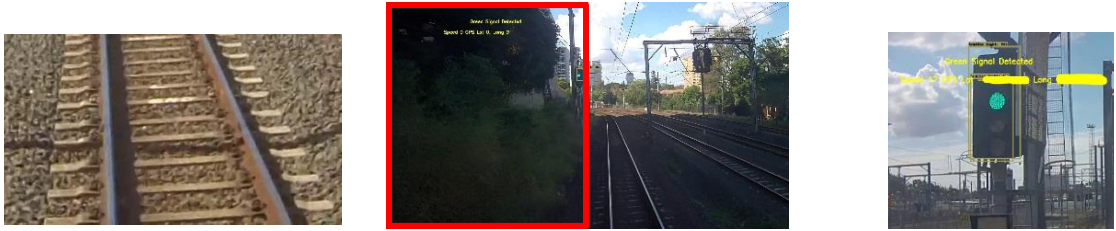

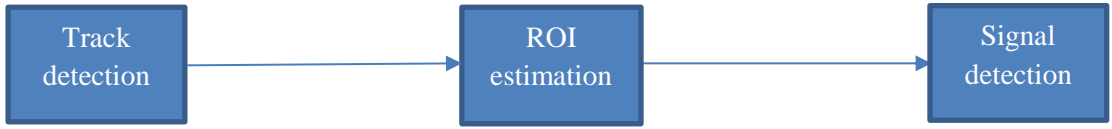

Figure 3 Track specific signal detection pipeline

### 3.4.1 TRACK DETECTION

*Edge detection based*
Track is prominent converging edge, this is a unique signature, which made edge detector the first choice for track detection. The image was filtered with Mexican hat filter and canny edge detection was applied. Hough's line transform was applied to obtain the prominent lines. If only one track is present, the lines corresponding to left and right part of the track can be consolidated by sorting the lines depending on their slope. If there are multiple tracks, they can be sorted depending on their slope as well as their distance from origin. It was observed that though this logic works well for the main track, it fails to detect the distant tracks by getting confused with shadows, slopes etc.

*Deep learning based holistically nested edge detection (HED)*
Deep learning based approach was attempted to cater to the shortcomings with conventional edge detection. This method leverages fully convolutional network and deeply-supervised nets to learn rich hierarchical representations that are important to approach the human ability to resolve the challenging ambiguity in edge and object boundary [4]. HED outperformed the conventional edge detection but couldn't distinguish stations, side rails from tracks as evident in Figure 4.

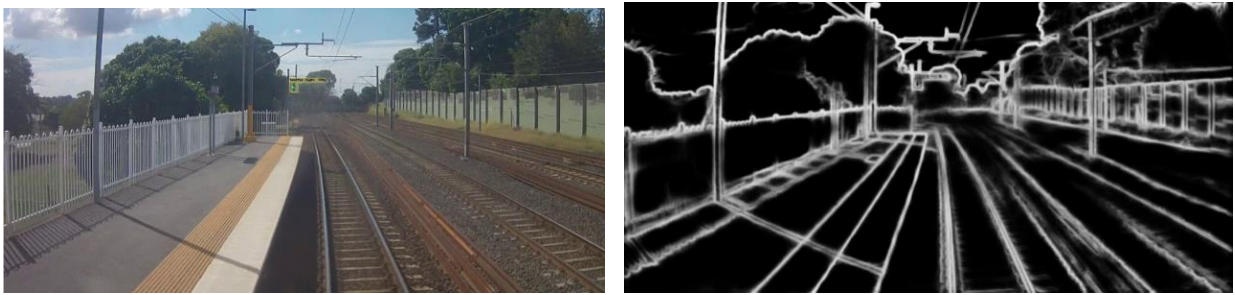

Figure 4 Edges detected using HED can't distinguish track, station, side rails

*Selected approach (simplified region based Convolutional Neural Network (CNN))*

Lot of information is lost in edge detection approach. Hence CNN based approach was attempted which is robust to presence of peripheral objects, vicinity to station and accommodates changes in environmental conditions. There are many architectures in literature which have been tuned to object detection task. [5][6] Region based CNN have the capability to generate region proposals and provide bounding boxes for the objects at the expense of model complexity and size. Additionally, the training images need to be annotated with the bounding boxes which is a labor-intensive process. If the study is confined to the lower quarter of the image, it is observed that the position, proportion, and alignment of track doesn't change a lot along the route as shown in Figure 2. Hence, region proposals can be snippets of lower quarter of the image which can be fed to simple CNN model in place of region based CNN. This fastens the training as well as evaluation process.

To train the CNN model an initial balanced dataset of 500 training images was made. The track images consisted of switches, double rails also. The negative class had other commonly objects present in images like ballast, vegetation, mileposts etc. Training an entire network is computationally expensive and impractical and hence transfer learning was used and inception V3 [7] model pretrained on Imagenet dataset was used. From the performance of ResNet [8] and Inception model in different cases in earlier works, it was observed that Inception model could learn well from a smaller dataset when the features weren't very complex as in the case of track detection. The training set comprised of tracks in sunny condition. Model was trained by tuning its last layer. It was seen that the same model performed well in cloudy condition as well, the same can be seen in the experimental results section.

### 3.4.2 ROI ESTIMATION

As mentioned earlier, the area where signal may be present is dependent on the position of main track w.r.t other tracks. Hence, once the tracks are identified, the region of interest needs to be isolated. It was observed that if the main track is on the left, the signal can be present anywhere on the left of the track. The signal is on the right for the rightmost track. And if the track is sandwiched between two tracks, the ROI is the region to the left of the main track till the next track on left as seen in Figure 2. Figure 5 gives the logic for ROI selection.

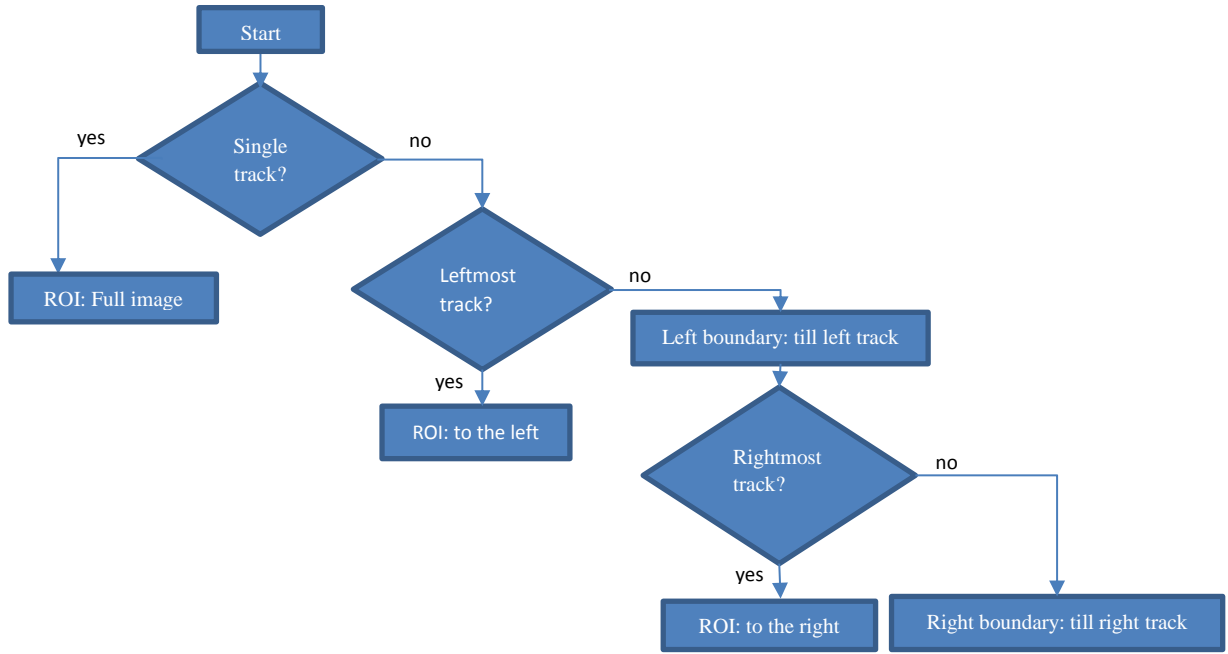

*Figure 5 Logic for Track specific ROI selection*

### 3.4.3 SIGNAL COLOR DETECTION

To detect the signal in the ROI selected, region based deep networks were selected since same frame could have multiple signals. Region based CNN models, pretrained on COCO dataset, which includes road signals as one of the categories was evaluated. To detect the signal color, red and green mask were applied on the image clipped based on the bounding box. The signals without color were ignored, which helped in reducing false positives. Faster R-CNN model [5] and Single Shot Detector (SSD) [9] models were evaluated. Results are shown in Table 1. SSD was seen to be faster than FRCNN but its detection accuracy was too low to be considered as a viable option. It was observed that glare in sunny days was reducing accuracy and it performed better in early morning videos or cloudy weather [10].

*Table 1 Comparison of SSD and Faster RCNN model for signal color detection*

|  | SSD | Faster RCNN |
| --- | --- | --- |
| Assets detected | 11/51 | 47/51 |
| Asset level accuracy (%) | 21.5 | 92.1 |
| Precision (%) | 98.4 | 98.6 |

# 4. EXPERIMENTS AND RESULTS

The GPU used to run these models is Titan X. The inception V3 model pretrained on Imagenet dataset was trained on dataset of 500 balanced data for two categories: track and not a track. This dataset was taken from a small portion of the route and was in sunny condition. Data was augmented by flipping. Shifting, shearing or rotation weren't applied as the scenario where the track is visible partially was to be avoided. An initial track specific signal selection (TSS) algorithm was made using the rules given in Figure 5 and Inception model trained.

Table 2 gives the initial results of the TSS algorithm. Even though the Inception model wasn't trained on cloudy/rainy condition, algorithm gives a good performance.

*Table 2 Results of the initial version of TSS algorithm for different environmental conditions*

| Environmental condition | Sunny | Rainy |
| --- | --- | --- |
| Distance covered (km) | 40 | 140 |
| Total Signals | 51 | 30 |
| Signals detected | 45 | 22 |
| Accuracy (%) | 88.2 | 73.3 |

Table 4. It was observed that the performance of this nascent TSS model was hampered in certain regions when a track was far away. The track wouldn't be visible on the lower quarter of the image, hence inception model wouldn't detect that track. But the track converges on the upper part of the image, hence it made the algorithm to select wrong ROI as shown in Figure 6. To cater to this problem, the height of the region proposal for inception model was increased by 25%. This resulted in an improvement in the accuracy by 60% for the specific case as seen in Figure 6 and Table 5. Figure 7 shows the cases where the track is far and cannot be captured by the inception ROI.

*Table 3 Confusion matrix for TSS performance for 129 signals in sunny condition*

| Actual\Predicted | Positive | Negative |
| --- | --- | --- |
| Positive | 120 | 9 |
| Negative | 17 | NA |

*Table 4 Error analysis for initial version of TSS algorithm*

|  | Side obstruction (railing/locomotive) | Curvy track | Station misidentified as track | Double rail/side rails | Side track is far (not in inception ROI) | Track Misdetection (grass, cemented surface/ditch) | Track not visible (dimness, glare, vegetation growth) |
|---|---|---|---|---|---|---|---|
| No. of misdetection | 4 | 6 | 4 | 1 | 17 | 3 | 6 |
| Time complexity to reduce error | difficult | medium | easy | medium | easy | easy | easy |

*Table 5 Reduction in misdetection with improved ROI*

| Misdetection by Inception V3 (Inception ROI 0.75h to h) | Misdetection by Inception V3 (Inception ROI 0.625h to h) |
|---|---|
| 17 | 7 |

h = height of the image

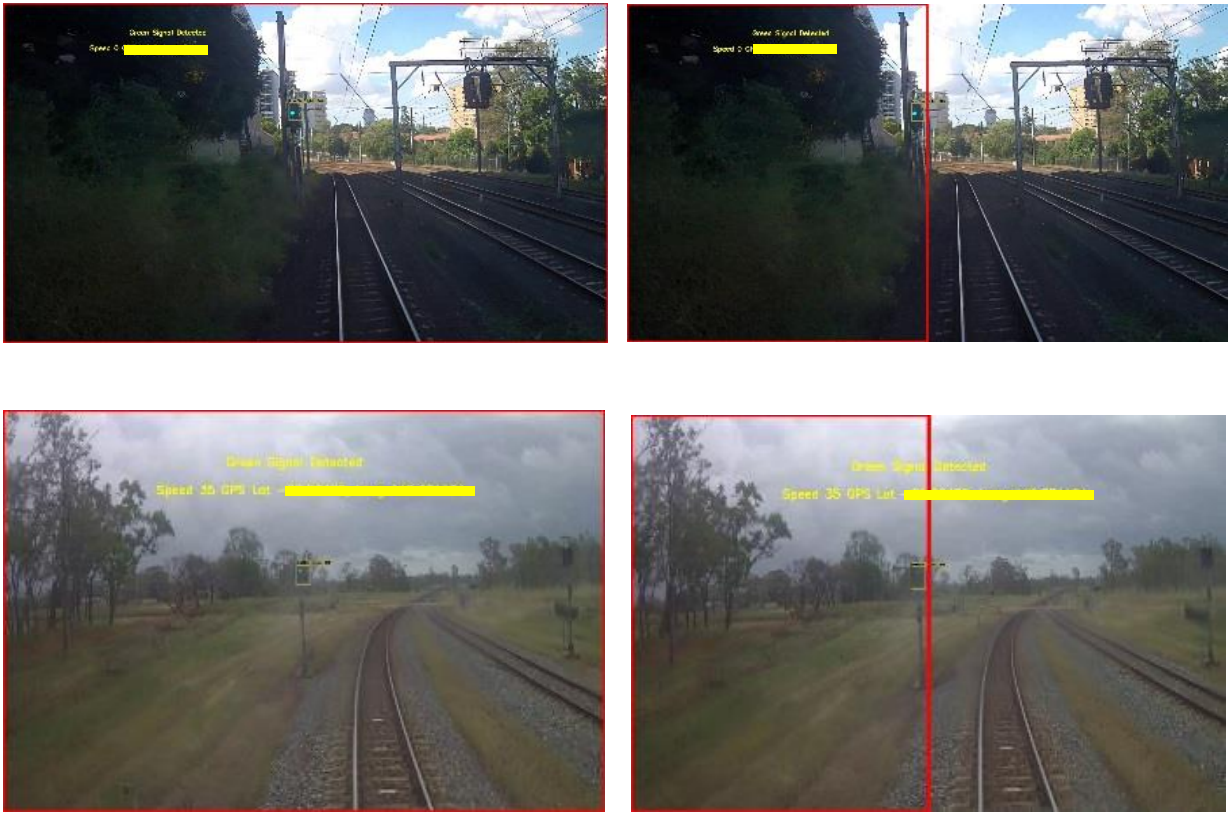

*Figure 6 Improvement in ROI selection (right) with ROI for inception model 0.625h to h from earlier case (Inception ROI 0.75h to h)*

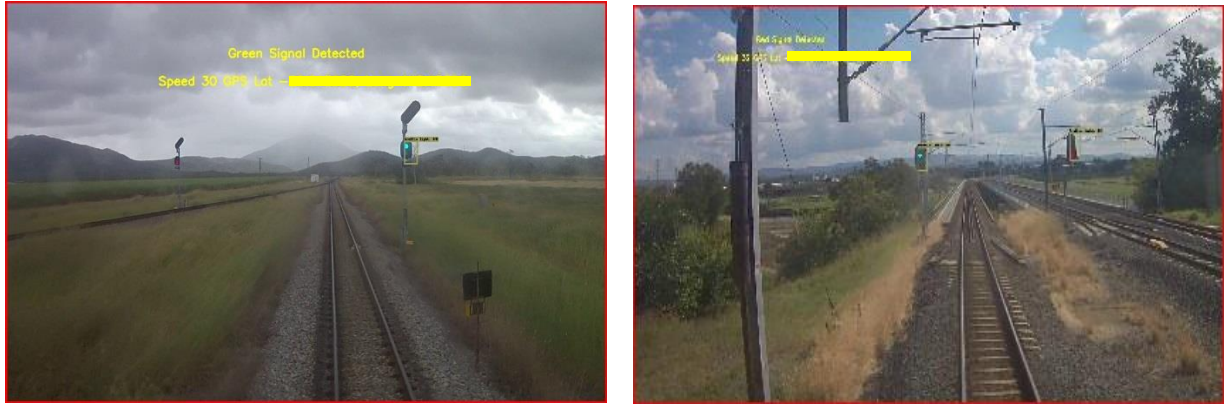

*Figure 7 Faraway tracks cannot be seen by inception model*

*Table 6 Error analysis for the TSS algorithm with improved Inception ROI*

|  | Side obstruction (railing/locomotive) | Curvy track | Station misidentified as track | Double rail/side rails | Side track is far (not in inception ROI) | Track Misdetection (grass, cemented surface/ditch) | Track not visible (dimness, glare, vegetation growth) |
|---|---|---|---|---|---|---|---|
| No. of misdetection (TSS_1) | 4 | 6 | 4 | 1 | 17 | 3 | 6 |
| No. of misdetection (TSS_2) | 4 | 6 | 4 | 1 | 7 | 3 | 6 |
| Time complexity to reduce error | Difficult | Medium | Easy | Medium | Difficult | Easy | Easy |

It can be seen from Table 6 that the next improvement can be obtained if the Inception algorithm can be finetuned to recognize the track better than earlier. To do the same, data was augmented for brightness. Special efforts were put to ensure the model doesn't confuse the ditch in between station, side railing as a track. Some of the examples are given in Figure 8.

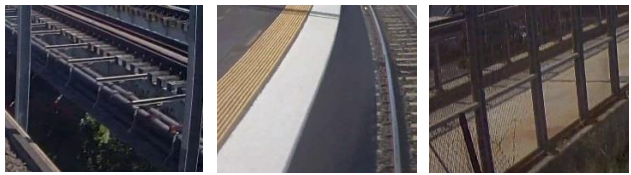
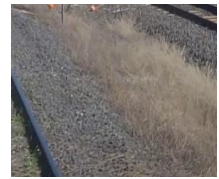

*Figure 8 Deceiving track false positives*

*Figure 9 Spacing between track mis-detected as track by Inception model*

It was also observed that during curves, the area in between two tracks was detected as a track which resulted in wrong ROI being selected. This was because two different rail were available from two tracks and the model understood it to be one single track as can be seen in Figure 8. Hence, the region fed to the inception was selectively kept 0.625h to h only for the edges. Table 7 gives the combined result from finetuning inception and selective change in inception ROI.

*Table 7 Error analysis with finetuned inception model*

|  | Side obstruction (railing/locomotive) | Curvy track | Station misidentified as track | Double rail/side rails | Side track is far (not in inception ROI) | Track Misdetection (grass, cemented surface/ditch) | Track not visible (dimness, glare, vegetation growth) |
|---|---|---|---|---|---|---|---|
| No. of misdetection (TSS_1) | 4 | 6 | 4 | 1 | 17 | 3 | 6 |
| No. of misdetection (TSS_2) | 4 | 6 | 4 | 1 | 7 | 3 | 6 |
| No. of misdetection (TSS_3) | 4 | 6 | 1 | 1 | 7 | 2 | 3 |

The analysis was done for a distance of 150km with a total of 247 signals. Table 8 gives the comparison of various TSS versions and an accuracy of 94.7% could be achieved. Confusion matrix for the final result is given in Table 9. Figure 10 gives some of the results with signals specific to track being selected in the ROI. There are certain cases were TSS can fail when the track is not visible like inside a tunnel or train occluding the track as can be seen in Figure 11

*Table 8 Comparison of signal detection performance of various TSS versions*

|  | TSS_1 | TSS_2 | TSS_3 |
|---|---|---|---|
| Distance covered (km) | 150 | 150 | 150 |
| Total Signals | 247 | 247 | 247 |
| Signals detected | 210 | 221 | 234 |
| Accuracy (%) | 85% | 89.4% | 94.7% |

*Table 9 Confusion matrix for final TSS version*

| Actual\Predicted | Positive | Negative |
|---|---|---|
| Positive | 234 | 15 |
| Negative | 24 | NA |

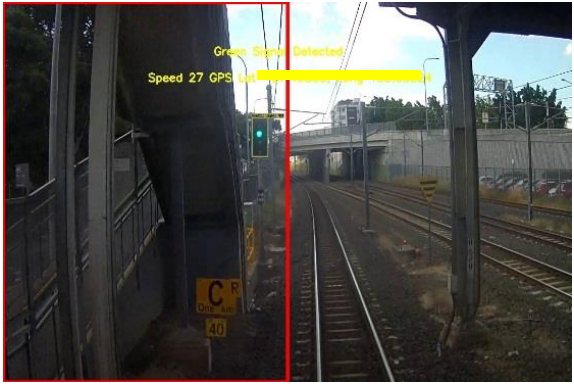
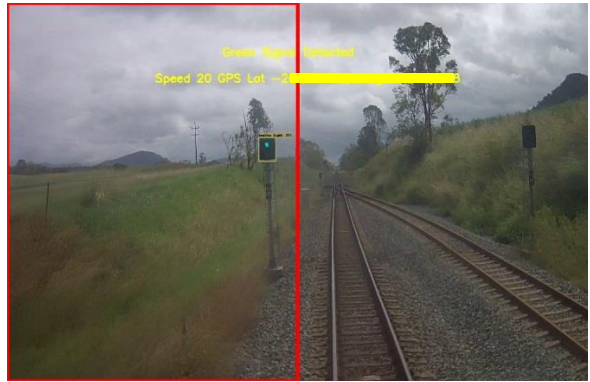
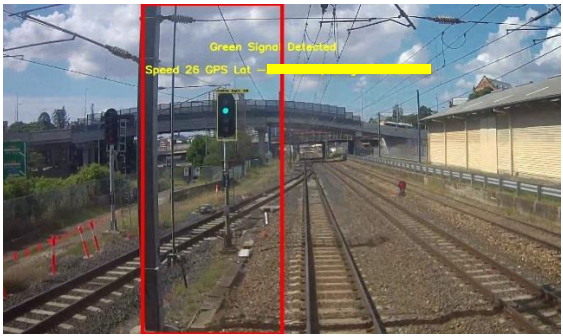
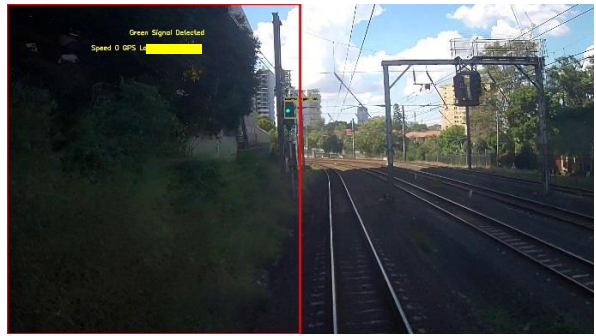

*Figure 10 Track specific signal selection results*

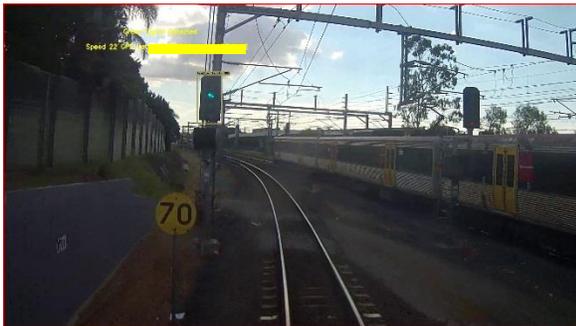
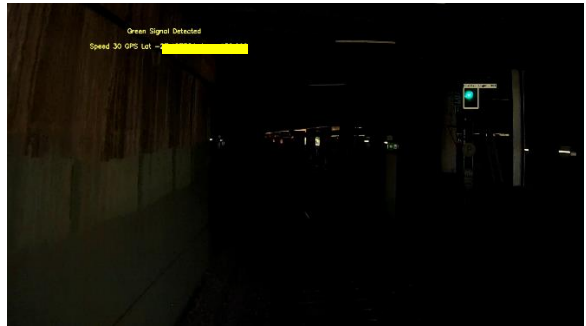

*Figure 11 Failure modes for TSS algorithm*

## 5. CONCLUSION

A track specific signal detection was implemented. Various approaches were conceptualized and tested for the same. Fixed ROI approach was found to have lot many failure modes. It was seen that the signal position was dependent on the presence and placement of other tracks. An object classification alternative was disregarded for being labor intensive. Track detection, ROI selection, signal detection was the pipeline chosen. For track detection, it was found that edge based weren't robust hence an inception model was trained for the same. It was trained for sunny condition and was found to work well even for rainy condition with an accuracy of 73%. Logic was developed for ROI selection based on relative placement of tracks. FRCNN model was used for signal detection and color detection was added on top of it. Error analysis was performed which propelled selective tuning of ROI fed to inception to include far away tracks without affecting the track detection efficiency, finetuning of inception to become robust to change in brightness and reduce misdetection of railing, station. An overall accuracy of 94.7% could be achieved on a route covering 150km with 247 signals.